\documentclass[twoside,preprint,11pt]{article}

\usepackage[hyperref]{jmlr2e}

\usepackage{amsmath,amssymb,amsfonts}
\usepackage{algorithmic}
\usepackage{graphicx}
\usepackage{textcomp}
\usepackage{booktabs}
\usepackage{longtable}

\usepackage{orcidlink}

\usepackage{marvosym}

\usepackage{lastpage}
\jmlrheading{}{2023}{}{}{}{}{Teemu Niskanen, Tuomo Sipola and Olli Väänänen}

\ShortHeadings{Recent Trends in Artificial Intelligence Technology}{Niskanen, Sipola and Väänänen}
\firstpageno{1}

\begin{document}

\title{Recent Trends in Artificial Intelligence Technology: A~Scoping Review}
\footnotetext[1]{
This research was supported in part by the \emph{Data for Utilisation – Leveraging Digitalisation Through Modern Artificial Intelligence Solutions and Cybersecurity} project and the \emph{coADDVA – ADDing VAlue by Computing in Manufacturing} project funded by the European Regional Development Fund (ERDF) and Recovery Assistance for Cohesion and the Territories of Europe (REACT-EU), under Grant A76982 and Grant A77973, and the \textit{Resilience of Modern Value Chains in a Sustainable Energy System} project, co-funded by the European Union and the Regional Council of Central Finland (grant J10052).}

\author{\name Teemu Niskanen \orcidlink{0009-0004-7615-2104}\AND
\name Tuomo Sipola \orcidlink{0000-0002-2354-0400} \email tuomo.sipola@jamk.fi\AND
\name Olli Väänänen \orcidlink{0000-0002-7211-7668} (\Letter) \email olli.vaananen@jamk.fi\\ 
\addr Institute of Information Technology\\
Jamk University of Applied Sciences\\
PO Box 207, FI-40101, Jyväskylä, Finland
}
\footnotetext[2]{\Letter\ Corresponding author: Olli Väänänen (e-mail: olli.vaananen@jamk.fi)}


\editor{Preprint submitted to arXiv.}

\maketitle

\begin{abstract}
Artificial intelligence is more ubiquitous in multiple domains. Smartphones, social media platforms, search engines, and autonomous vehicles are just a few examples of applications that utilize artificial intelligence technologies to 
enhance their performance. This study carries out a scoping review of the current state-of-the-art artificial intelligence technologies following the Preferred Reporting Items for Systematic Reviews and Meta-Analyses (PRISMA) framework. The goal was to find the most advanced technologies used in different domains
of artificial intelligence technology research. Three recognized journals were used from artificial intelligence and machine learning domain: Journal of Artificial Intelligence Research, Journal of Machine Learning Research, and Machine Learning, and 
articles published in 2022 were observed. Certain qualifications were laid for the technological solutions: the technology must be tested against comparable solutions, commonly approved or otherwise well justified datasets must be used while applying, 
and results must show improvements against comparable solutions. One of the most important parts of the technology development appeared to be how to process and exploit the data gathered from multiple sources. 
The data can be highly unstructured, and the technological solution should be able to utilize the data with minimum manual work from humans. The results of this review indicate that creating labeled datasets is very laborious, and solutions exploiting unsupervised or semi-supervised learning technologies
are more and more researched. The learning algorithms should be able to be updated efficiently, and predictions should be interpretable. Using artificial intelligence technologies in real-world applications, safety and explainable predictions are mandatory to consider
before mass adoption can occur.
\end{abstract}



\section{Introduction}
Artificial intelligence (AI) technologies try to solve problems using different approaches in various domains. Recent years have shown remarkable growth in business adoption of AI~\cite{iansiti2020}. The AI approaches can be categorized based on the task the technology tries to solve and the data it uses
while solving the task. This study concerns some of such technological approaches as listed next. Natural language processing is an approach that tries to process human language~\citep{NLP}. Natural language processing can be used to solve different tasks such as text classification~\citep{fake}, text summarization~\citep{succinct}, and text translation~\citep{simultaneous,context_aware}. It can be used to separate fake news from real ones~\citep{fake} or to provide succinct 
description of the events~\citep{succinct}. Computer vision is an approach that tries to process and utilize images and videos~\citep{vision}. Computer vision can be utilized in tasks such as identifying artificially generated images from real images~\citep{artificial_images} or in predicting events such as disease outbreaks~\citep{spatio}.
Reinforcement learning is an approach where the artificial intelligent agent interacts with an unknown environment~\citep{RL}. While the agent interacts with the environment the state of the environment changes and the agent receives rewards based on the state~\citep{RL}. Reinforcement learning can be
utilized in tasks such as imperfect-information game~\citep{starcraft},~\citep{interpretable}, and imitating animal or human cognition~\citep{physical}. Motion planning is an approach that tries to model motions of autonomously operating systems such as self-driving cars~\citep{CCTMP}. Motion planning can be utilized while developing autonomous vehicles~\citep{CCTMP}.

AI, while being a promising technology, has societal and ethical implications. Individual level autonomy and even human dignity could become issues with non-human agents. 
More traditionally, privacy and data protection also extend to the world of AI, especially because huge volumes of data about humans is being used. 
In addition, fairness, equity and diversity could become a problem with the patterns learned from training data. 
Autonomous agents raise the question of responsibility and accountability, and the transparency of these systems is not always clear, sometimes stemming from the technological structure~\citep{vecnicalujevic2020}. 
Ethical impact of AI can be considered in various domains. 
Facts such as the complexity of other systems and AI's interactions with them could obscure the real impact of AI. 
Data security, privacy and misuse could be a new problem. 
As mentioned above, transparency and reliability are also concerns. In addition, accessibility is also topical. 
Finally, training data bias can also be identified as a challenge, and the recommendations given by the AI systems could include misleading information~\citep{zhang2023}. 

We use a hierarchical model with three levels: high-level branches, tasks, and finally, technological solutions. In this scoping review, technological solutions were found for 21 different tasks. These tasks were grouped into five branches: Natural language processing, Computer vision, Robotics and motion, Reinforcement learning, and Others. Natural language processing contains tasks involving textual data and Computer vision tasks involving visual data. Robotics and motion contains tasks that involve motion planning and trajectory
prediction of autonomous vehicles. Reinforcement learning includes tasks that exploit reinforcement learning agents while interacting with unknown environments. The branch Others contains tasks that cannot be distinctly classified into any of the other branches. Under each branch, different tasks and problems are introduced with a technological solution that has shown state-of-the-art performance when tested against comparable solutions. A task can include more than one problem, and problems are provided with technological solutions. 

Several published scoping reviews can be found, and many of these publications concentrate on a certain field. Health care domain especially is widely reviewed, and there are overall reviews, which highlight the use of AI in patient-provider encounters, support human decision-making and effectiveness of interventions~\citep{health}. In this field, there are also aspect specific reviews such as predicting cardiac arrest~\citep{cardiac} or using artificial intelligence for MRI diagnosis~\citep{MRI}. Reviews can also be found from construction field~\citep{construction}, education~\citep{education}, and automotive industry~\citep{car}.
Additionally, surveys concerning certain tasks of artificial intelligence technologies such as generating language~\citep{language_survey} or creating intrinsically motivated reinforcement learning agents~\citep{intrinsic} can be found.
There seems to be a lack of reviews mapping technological solutions used while solving different real-world tasks and problems. 
All the found reviews are focused on a single domain. Therefore, we aim to present a top-down hierarchical view of the latest trend in AI technology, focusing on applied results and real use cases.

The artificial intelligence technologies are developing rapidly, and this scoping review was conducted to find the most recent and advanced artificial intelligence technological solutions for solving various tasks. The purpose was to systematically map the most advanced and proven 
technological solutions developed and used in the year 2022. This scoping review was conducted to provide a wide overview of solutions, and the review does not concentrate on any particular task or domain.

The structure of the paper is as follows: First we present the methodology behind the scoping review. Then the results for each branch are presented. A discussion section summarizes the main findings, and finally, the conclusion describes some recommendations and future avenues. A list of abbreviations is provided as an appendix. 

\section{Methods}
This study was conducted using methodology based on the scoping review~\citep{arksey2005scoping} extension provided by Preferred Reporting Items for Systematic Reviews and Meta-Analyses (PRISMA) \citep{PRISMA}. PRISMA provides an evidence-based minimum set checklist with explanations of reporting items to include in a scoping review. 
In this scoping review, scientific publications were used from artificial intelligence and machine learning domains. In order to focus on recent research and to limit the total number of studies, articles that were published in 2022 were selected. 
Flowchart of the process is shown in Fig 1, according to the PRISMA framework. 

\begin{figure*}[t]
    \centering
    \includegraphics[width=15.5cm]{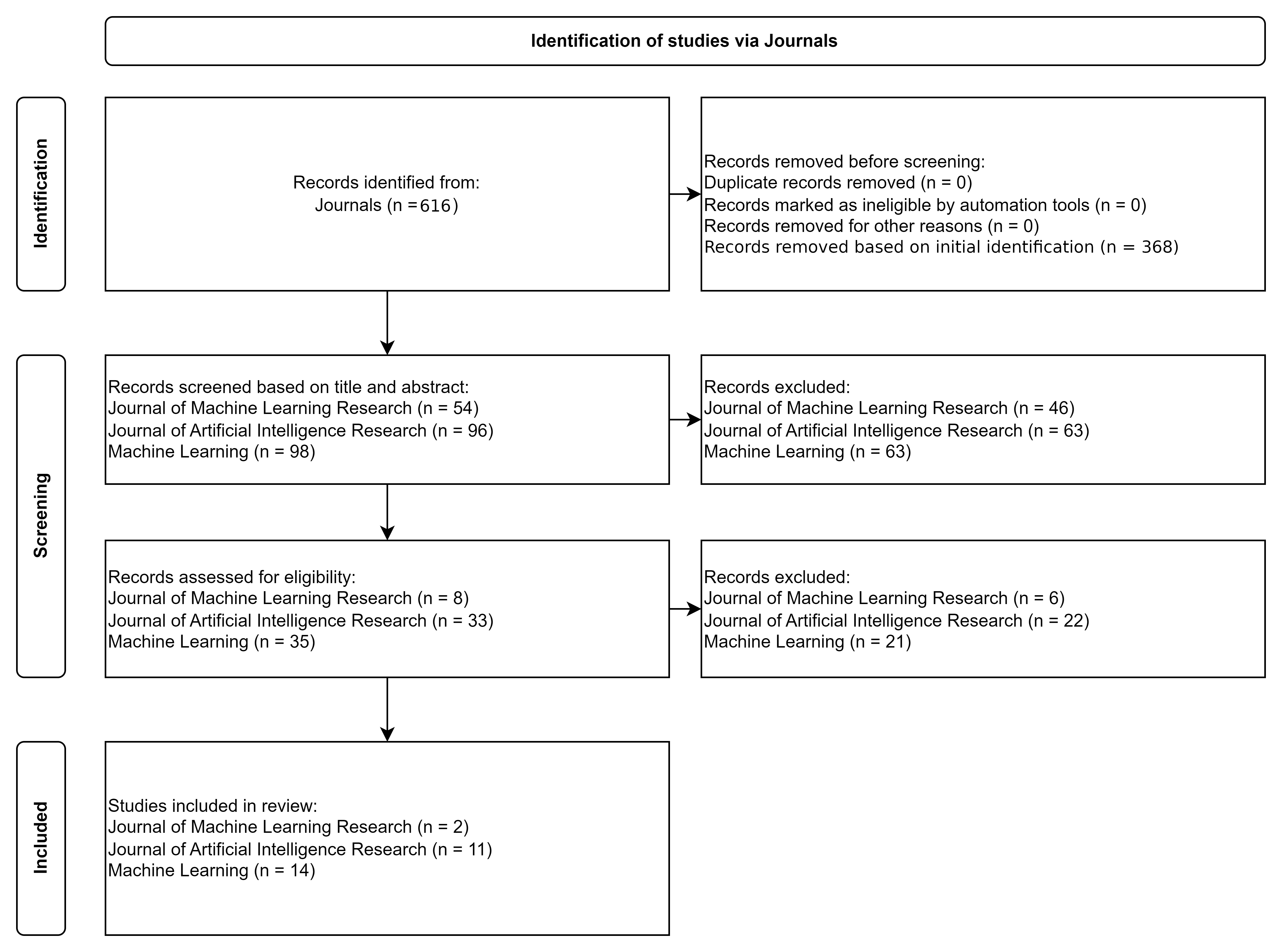}
    \caption{Flowchart of identification of studies via Journals.}
    \label{Fig 1}
\end{figure*}

\subsection{Journal selection}

Three leading journals according to certain national ranking lists were used: 
Journal of Artificial Intelligence Research (ISSN 1076-9757), Journal of Machine Learning Research (ISSN 1533-7928), and Machine Learning (ISSN 1573-0565). These journals were selected based on the Finnish Publication Forum~\citep{publication} rating and the relevance to the domain. 
Each journal was required to have level 3 (which is the highest level), classification and a focus on artificial intelligence or machine learning technology. 
It should be noted that all three journals are at the highest level (2) in the Norwegian Register for Scientific Journals, Series and Publisher~\cite{norway}.
In addition, Artificial Intelligence (ISSN 0004-3702) could be considered. 
However, the three journals were seen as sufficiently representing the field within the constraints of reasonable time. All volumes and issues published in 2022 in these journals were included
in the identification process. 

In summary, the selection criteria for the journals were: 

\begin{itemize}
\item domain of the journal: artificial intelligence or machine learning,
\item published in 2022,
\item highest level in Finnish and Norwegian national journal ranking.
\end{itemize}

\subsection{Initial identification}

There was no search phrase used, because all articles from the selected journals published during 2022 were included in the identification phase. 
The articles were downloaded from the websites of the journals themselves. 
In total, the three journals published 616 articles during the year 2022. 
Each article was separately inspected, and only articles that had their first publication date in 2022 were qualified. 
In total, 248 articles were initially identified by title and abstract to select potentially relevant ones. 
The following criteria were used to identify the articles:

\begin{itemize}
\item AI methodology is applied to a real problem,
\item includes a proposed solution,
\item base technology is AI-based.
\end{itemize}

There were no identified duplicate records, and all the records could be read by the researchers. 

\subsection{Screening}

From these 248 articles, 76 articles were evaluated based on title, abstract and preliminary inspection of the full text to screen for relevant publications. 
The articles were included based on the technological solutions used in them: the selected technology must be applied against comparable solutions, commonly approved or otherwise well justified datasets must be used while applying, and the results must show improvements in some areas against comparable solutions.
In order to qualify, the article should include comparison between methods. 

\subsection{Inclusion in the study}

From these 76 articles, 27 articles were selected for charting. 
A data charting form was created to determine the data to be extracted. The eligibility criteria for inclusion in charting were: 

\begin{itemize}
\item AI methodology is applied to a real problem,
\item includes a proposed solution, 
\item base technology is AI-based, 
\item real datasets are used while testing the solution, 
\item solutions that were compared against others while evaluating the performance, 
\item testing results are shown. 
\end{itemize}

Excluded articles covered theoretical topics, or did not include an applied angle to the problem to be solved. The articles were categorized based on the task and the problem that the technology tried to solve.
Each of the 27 articles were summarized, and each such summary document included:

\begin{itemize}
\item title of the article (e.g, Agent-Based Modeling for Predicting Pedestrian Trajectories Around an Autonomous Vehicle),
\item real-world problem to be solved (e.g., predicting pedestrian trajectories)
\item task type (e.g., trajectory prediction),
\item technology family (e.g., Agent-Based Modeling (ABM)),
\item AI model used in the study (e.g., expert pedestrian model),
\item technologies used (e.g., Agent-Based Modeling (ABM): social force modeling (SFM) + decision model),
\item datasets (e.g., DUT, CITR, Nantes),
\item compared to (e.g., standard SFM),
\item a few paragraphs summarizing the article,
\item clarifying images lifted from the article,
\item full reference.
\end{itemize}

\section{Results from the scoping review}
Based on the results from the scoping review the artificial intelligence technologies are divided into five branches: Natural language processing, Computer vision, Robotics and motion, Reinforcement learning, and Others. These branches were selected to separate the technological solutions based on the tasks and the problems the technology tries to solve.
In this study Natural language processing includes tasks that process textual data, Computer vision includes tasks that process visual data, and Reinforcement learning includes tasks that exploits reinforcement learning agents. Robotics and motion includes tasks that are related to the motion of autonomous 
systems. Trajectory prediction is also included in this branch. The branch called Others includes tasks that cannot be distinctly inserted in any of the aforementioned branches. However, these tasks may include some aspects of the other branches.
Under each branch, different tasks and problems are introduced with most recently proposed solutions. This division was made based on the observation obtained from the scoping review, where differences in tasks and problems could be highlighted based on the data that is processed and generated. Also, reinforcement learning agents can be distinctly separated as 
their own branch.

In the following sections each branch is introduced with tasks and problems. The branches are presented in the following order: Natural language processing, Computer vision, Robotics and motion, Reinforcement learning, and Others. Each branch is also illustrated with a diagram.

\section{Natural language processing}

Natural language processing (NLP) is a branch of artificial intelligence which tries to process human language. Many useful real-world applications rely on NLP such as text translation and text summarization. NLP can
be utilized in tasks that could be impractical or very laborious using human workers. NLP widely uses traditional machine learning and neural network solutions. Using deep learning with NLP, state-of-the-art
results have been obtained~\citep{NLP}. In this study tasks and problems that process textual data are included in this branch. The structure of the branch is illustrated in Fig 2.  

\begin{figure*}[t]
  \centering
  \includegraphics[width=\linewidth]{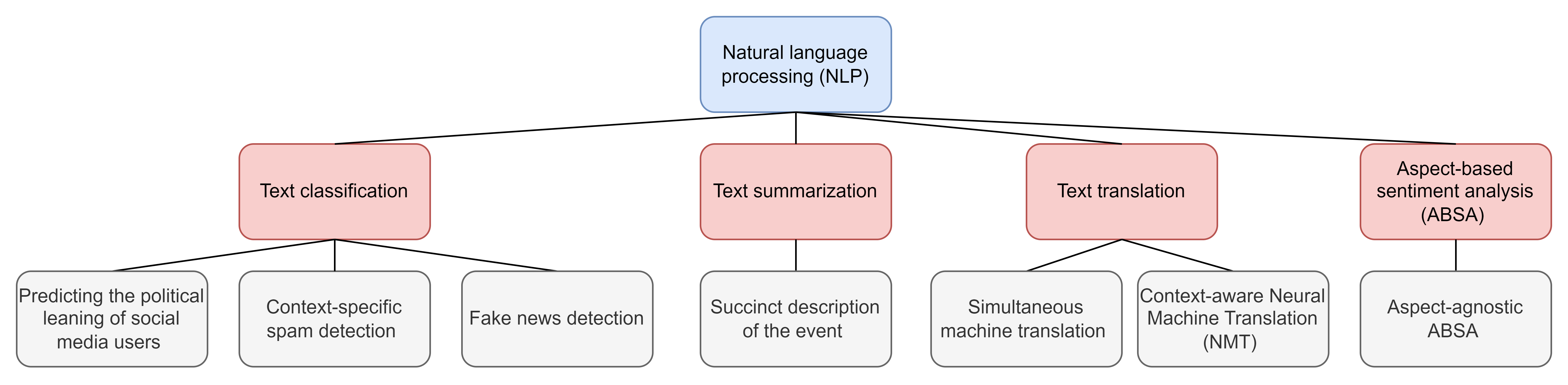}
  \caption{Structure of the Natural language processing branch.}
  \label{Fig 2}
\end{figure*}

\subsection{Text classification}

Text classification can be used in tasks such as fake news detection~\citep{fake}, spam detection~\citep{context} and predicting
political leaning~\citep{political_leaning}. Supervised learning algorithms are commonly used in text classification. The performance of
these algorithms has been improved by using different methods such as advanced feature extractors~\citep{context}. However, 
supervised learning algorithms have a major shortcoming due to the need of labeled datasets~\citep{fake}. New methods implementing
unsupervised learning and semi-supervised learning algorithms are extensively researched~\citep{fake},~\citep{political_leaning}.

\subsubsection{Fake news detection}

Most research on detecting fake news using text analysis relies on binary classification, which 
requires a large and balanced set of labeled news articles. Here, balanced means that both classes are presented by equal number samples from each class. Considering balanced scenarios, datasets 
that are used for evaluating performance of these algorithms, usually do not reflect the real world. 
Therefore, there is a need for semi-supervised learning approaches that require less labeled news 
and are more effective in a variety of news datasets~\citep{fake}.

One-Class Learning (OCL) algorithms have been extensively used in text classification. OCL algorithms use only a single class while learning 
and the designated class is usually the subject of interest. The classifier then identifies samples belonging to that one class, while classifying everything else as not belonging to it. Most promising performances have been obtained by using k-Means, k-Nearest Neighbors 
Density-based (k-NND), and One-Class Support Vector Machine (OCSVM). The performance of these algorithms has been proven in literature for 
multidimensional data classification, but using these algorithms in the context of news classification has 
some limitations because used datasets and given parameters have a notable impact on performance~\citep{fake}.

Positive and unlabeled learning (PUL) algorithms generate a set of data including labeled 
interest and non-interest data, and unlabeled data. After the data has been generated, inductive 
semi-supervised learning algorithm or transductive semi-supervised learning algorithm will be used 
for labeling the unlabeled data. Interest data class needs to be labeled manually, but PUL algorithms 
reduce the need for labeling interest class data, and only a small amount of data needs to be labeled. 
Regarding text classification, most promising results from PUL algorithms have been obtained by 
using Rocchio Support Vector Machine (RC-SVM). RC-SVM uses Rochhio for outlier extraction from unlabeled data and these outliers 
will be used as non-interest data. After extraction Support Vector Machine (SVM) will be used 
iteratively for classification~\citep{fake}.

\citet{fake} introduced Positive and Unlabeled learning by Label Propagation (PU-LP) algorithm, which is
a PUL algorithm with label propagation extension. PU-LP algorithm also reduces labeling efforts 
and in addition, PU-LP algorithm uses unlabeled data to extract relevant information. Relevant 
information will assist the algorithm to improve classification. Label propagation algorithm will use 
the extracted information in labeling the unlabeled data. Based on the results PU-LP outperforms PUL algorithms, OCL algorithms and binary algorithms
in most of the test scenarios. 

\subsubsection{Context-specific spam detection}

Content that can be considered as spam is rapidly increasing, and it is a major problem of the internet. 
In social media, advertising is a commonly used and approved method. Usually if advertisement is not 
related to the context in question, it is seen as spam. This can be seen as an advertisement mismatch 
rather than traditional spamming. Conventional spam filters categorizing all advertisement as 
spam are not the solution because they prohibit all kinds of advertisement. Contextual 
understanding is required to separate advertisement mismatch from traditional spam. Most common state-of-the-art solutions for spam detection use content information and user information. A shortcoming of 
these solutions is that user information is not always available. Most promising results have been 
obtained by using models based on Random Forest~\citep{context}.

Context-specific spam has been researched by~\citet{context}. The study used tweet content from Twitter. The purpose was to separate traditional spam from 
context-specific spam. Context-specific spam is a post that is irrelevant for the context in question. The goal of the study 
was to research automated solutions for irrelevant content removal from a domain using Twitter data. 
The focus was on low resource setting where limited training data is available and training data is 
imbalanced. Classical machine learning algorithms used in spam detection such as Random Forest and Support Vector Machine (SVM) were compared against each other and against a neural network solution. 
Feature extractors such as Bag-of-Words and Word Embedding were used with classical machine learning algorithms.
The neural network solution structure contained fine-tuned pre-trained language model Bidirectional Encoder Representations from Transformers (BERT) and a single layer neural network. 
Based on the results the best performer with single domain dataset is neural network with pre-trained 
language model. Considering classic machine learning models, logistic regression and random forest 
were the best performers. With multi-domain dataset, classic models were more robust than the ones based on neural network. 

\subsubsection{Predicting the political leaning of social media users}

Nowadays people are consuming political content from social media platforms. This trend has led to 
a situation where it is mandatory for politicians to have a digital campaign. Predicting political 
leaning of social media users has drawn much attention in recent years. Based on political leanings, 
different analyses can be conducted, such as forecasting the outcome of the election or measuring online 
polarization. Most of the studies have concentrated on the content of messages or social networks, 
although there are few exceptions using both the content and the networks. A flaw for network-based 
approaches is that they usually make an assumption that similarly thinking users act together and follow each other, and that can be violated. Also, they need a 
huge amount of data, which is not often available. Content-based approaches usually rely on 
supervised learning and require natural language processing. Using supervised learning has shown 
some major shortcomings such as limited availability of datasets and lack of generalizability~\citep{political_leaning}.

An unsupervised learning content-based solution was 
introduced by Fagni and Cresci \cite{political_leaning} for predicting political leaning. The main goal was to predict the preferred political party 
and the political pole (left-right ideology) of the Twitter users. In the proposed solution a neural network was used as a classifier, 
Uniform Manifold Approximation and Projection (UMAP) was used for feature reduction, and K-Means, GaussianMixture, and MeanShift algorithms were 
used for data clustering. Three different methods were created and these methods were compared against each other and different unsupervised learning, supervised learning and 
semi-supervised learning solutions. Based on the results, the method called ``Parties enriched + clustering'' was the best performer between proposed methods. This method outperforms other unsupervised learning solutions
but cannot outperfrom advanced supervised learning and semi-supervised learning solutions. However, the difference in 
performance was not significant, and it must be noted that the solutions that are not based on unsupervised learning require labeled data to perform.

\subsection{Text summarization}

The vast amount of information from different sources such as numerous media providers has caused 
information overload. Summarization of events from the sea of information is a great value for news 
systems and search engines to provide their users a quick perception of hot topics. In addition, extractive, abstract and hybrid methods can be used for summarization. Different variations
of Multi-document Event Summarization (MES) frameworks have been researched to produce succinct
descriptions of the events~\citep{succinct}.

\subsubsection{Succinct description of the event}

MES is an event summarization task. While conventional multi-document 
summarization tries to generate a summary containing multiple sentences from a set of documents, 
MES tries to summarize the core event in few words. The purpose of MES is to produce a query-level 
summary, which is essential for example for search engines~\citep{succinct}. 

In generation and summarization tasks encoder-decoder neural network models have shown robust 
representation capabilities. However, capturing relation in different documents and eliminating 
redundancies are major shortcomings while operating settings with multiple documents. 
Graph-based methods have presented good performance while handling multiple documents. However, 
syntactic information loss is a main flaw with graph-based methods~\citep{succinct}.

\citet{succinct} introduced a MES framework called Event-Pg. 
The Event-Pg contains two different phases; the first phase an event identification phase, and the second phase a 
generation phase. In the event identification phase events are detected on sentence level and on graph-level. 
In the generation phase an event-aware pointer generator is used for event sequence generation.
Based on the results, Event-Pg was very effective against baseline algorithms such as mBart. 

\subsection{Text translation}

In conventional machine translation (MT) complete sentences are translated into another language~\citep{simultaneous}.
Different methods have been researched to improve the quality of translation such as exploiting the contextual information~\citep{context_aware}.
Real-time translation, which tries to make translation based on partial data, has also been researched~\citep{simultaneous}.   

\subsubsection{Simultaneous machine translation}

Simultaneous MT is a task where an algorithm tries to translate speech in real-time. 
Simultaneous MT is constantly balancing between context accumulation and producing 
translation. Where conventional MT process complete sentences, simultaneous MT tries to figure 
out how much data it needs to produce a partial translation. Producing translation in real-time has 
been explored by using for example neural MT (NMT)~\citep{simultaneous}.

Multimodal machine translation (MMT) framework tries to improve translation quality by using 
additional sources such as images or videos as additional context. Several studies have been conducted 
to explore the benefits of grounding language with visual content in simultaneous MMT, for example 
visual features have been integrated into coders of recurrent MMT architecture. Most of the research considering simultaneous MMT 
relies on rule-based strategies using Multi30k dataset for recurrent MMT models~\citep{simultaneous}.

A Transformer-based simultaneous MMT model is introduced by~\citet{simultaneous}. The model includes a novel approach by supervising alignments of the source 
language representations and the image regions. The study indicates that supervised attention significantly improves the performance 
of simultaneous MMT. Based on the results the proposed solution performs slightly better than state-of-the-art simultaneous MMT models from \citet{Imankulova} and \citet{Caglayan}.

\subsubsection{Context-aware Neural Machine Translation (NMT)}

Neural Machine Translation (NMT) has shown success on translating sequences between languages. 
Recent studies have tried to leverage contextual information with NMT. In these cases, contextual 
information is added as an additional input. These inputs themselves are not translated but rather used as 
an assistance while translating desirable sentences. The main issue with context-aware translation has 
been how to effectively process the contextual information. A pre-trained method such as BERT 
has been used with conventional NMT, but there is a lack of studies trying to apply pre-trained 
methods to context-aware NMT~\citep{context_aware}.

\citet{context_aware} has studied context-aware NMT using 
pre-trained language model BERT to encode contextual information and receive contextual features. 
Three approaches were studied to aggregate the contextual features: Concatenation mode (C-
mode), Flat mode (F-mode), and Hierarchical mode (H-mode). In the C-mode, contextual sentences are concatenated and fed to BERT. In the F-mode, 
contextual sentences are independently fed to BERT and the outputs are concatenated. In the H-mode, 
contextual features are aggregated hierarchically. The H-mode uses contextual features produced by the F-mode
to further process them with a world-level and a sentence-level attention model. Based on the results the C-mode outperforms other proposed solutions in 
every task. The C-mode was compared against multiple baseline solutions, such as G-Transformer and BERT-NMT, which the C-mode outperformed considering BLEU score.

\subsection{Aspect-based sentiment analysis (ABSA)}

Aspect-based sentiment analysis (ABSA) tries to utilize aspects while trying to conclude the 
sentiment in sentences. This can be used for example when analyzing comments on the internet. 
Aspect term sentiment analysis (ATSA) and Aspect category sentiment analysis (ACSA) can be included in ABSA.
Aspect-agnostic methods have been researched with ABSA to improve the performance of standard context encoders~\citep{aspect}.

\subsubsection{Aspect-agnostic ABSA}

Long Short-Term Memory networks (LSTM), Graph Convolutional Networks (GCN) and 
pre-trained BERT are all used in ABSA models as a context encoder. LSTM is used as a sequence-based context encoder, GCN as a graph-based context encoder, 
and BERT as a pre-trained context encoder. They generate hidden states which are fed to the next module~\citep{aspect}.

According to~\citet{aspect}, in these hidden states aspect-specific information might be discarded and aspect-irrelevant information might be retained. The study 
introduced three context encoders with aspect aware (AA) mechanism. The encoders were based on 
LSTM, GCN and BERT. The results show the proposed AA mechanism improves the performance of the vanilla versions of the encoders.

\section{Computer vision}

Computer vision considers a wide range of applications which have become a part of our day-to-day life. Smartphones and photo sharing applications utilize the capabilities of computer vision while performing tasks. 
The development in deep learning techniques has enabled the implementation of more and more complex systems~\citep{vision}. In this study, tasks and problems processing visual data are included in this branch. The structure of the branch is illustrated in Fig 3. 

\begin{figure*}[t]
  \centering
  \includegraphics[width=\textwidth]{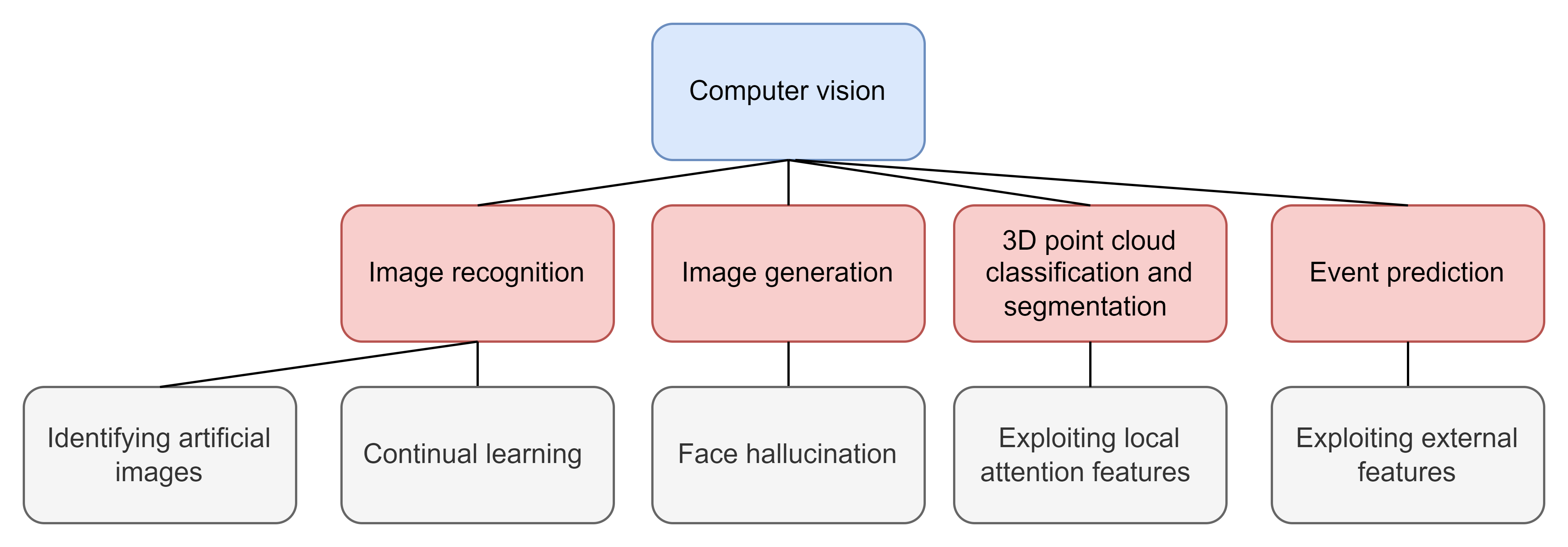}
  \caption{Structure of the Computer vision branch.}
  \label{Fig 3}
\end{figure*}

\subsection{Image recognition}

With the development of generative models, it has become more crucial that artificial images can be separated from real images. Algorithms must be taught to distinguish the difference between these images and one promising
solution is to use wavelets while processing images~\citep{artificial_images}. Another major challenge for image recognition algorithms is how to make the learning continuous. Most solutions require retraining of the whole model 
if new data is applied. Solutions have been introduced which only require certain module in the framework to be retrained, which makes the insertion of new data much more efficient~\citep{continual}.

\subsubsection{Identifying artificial images}

Generative adversarial networks (GAN) can be used for beneficial purposes such as generating images 
from text or extracting representations from data. GANs can also be used in malicious acts e.g., 
generating fake images with deception purposes. These so called deepfakes could cause a threat to 
integrity of knowledge. This threat should be confronted with technologies that can separate a real 
image from a fake one. Besides GAN, diffusion probabilistic models and score-based generative models have shown high-quality results as 
generating images~\citep{artificial_images}.

Deepfake detectors can be separated in two categories working in different domains: frequency 
domain and spatial or pixel domain. In the frequency domain, for example discrete cosine transform 
(DCT) and Fourier feature based detectors provides frequency space information from images but 
spatial relations are missing. With extracted information, classifying algorithm is trained. In latter 
category convolutional neural networks (CNN) are usually trained with raw images to identify 
diverse architectures from GAN. Also, denoising-filters, mean operations and computing 
convolutional traces have been applied~\citep{artificial_images}.

Wavelets were first introduced in applied mathematics, but they can also be applied in image 
processing and analysis. In recent research wavelets have been used within pooling layers in neural networks whereas 
previously they were used as input features or in scatter nets as early layers~\citep{artificial_images}.

\citet{artificial_images} conducted an experiment where 
two images were compared; one from Flickr Faces High Quality (FFHQ) dataset and the other one generated 
by StyleGAN. The purpose was to investigate the coefficients of the level 3 Haar wavelet packet 
transform. The absolute values were scaled using the natural logarithm (ln). Three aspects were 
compared: mean wavelet packets, standard deviations, and difference in mean wavelet packets and 
standard deviations. Significant differences between the two images were discovered. 
This difference was utilized, and three network architectures were created: Wavelet-Packet, Fourier and Pixel, and Fusion. Different wavelet packets were used with 
various degrees. Based on the results wavelet packets improved the performance of the classifier. 

\subsubsection{Continual learning}

Models integrating language and vision can be referred to as multimodal representation learning 
algorithms. The principle of these models is to ground the meaning of language with visual properties, which 
is crucial to artificial agents that interact with 
environment. Impressive results have been achieved by using Contrastive Language-Image Pre-Training (CLIP).
CLIP uses contrastive learning with images and their captions. Dot product is 
calculated for embedded images and texts. Based on calculations similarities are detected. CLIP has 
shown significant results in zero-shot learning. In contrast to conventional image classification 
algorithms using predefined category sets there is no limit for the amount of categories CLIP 
can use. CLIP has also proven more robust considering noise and variations compared to 
conventional image classification algorithms~\citep{continual}.

When acting in a dynamic environment it is crucial that the model can be updated, and the learning is 
continuous. The aforementioned algorithm CLIP cannot be updated without retraining the entire model. 
This is a major flaw considering its use in real-world applications. In real-world situations, for 
example in basic conversations between two people, the meaning of language may differ depending on 
participants. Models should be able to adjust themselves while interacting with environment~\citep{continual}.

\citet{continual} introduced a solution called Continual Learning of Language 
Grounding from Language-Image Embeddings (CoLLIE) to tackle the continuous 
problem. CoLLIE relies on CLIP as a foundation model, and the updates are created using a separate 
transformation model. Whereas in few-shot learning, for each concept a new model must be trained, in 
CoLLIE only the transformation model needs to be updated. When new concepts are learned, 
the transformation model adjusts text embeddings. Dimensions that are not misaligned stay 
unaffected. Based on the results, CoLLie appears to be sample efficient, and it can generalize. 
It was concluded that with CoLLIE a new benchmark was established for future studies.

\subsection{Image generation}

Most of the face recognition algorithms perform poorly when low-resolution images are used. One solution for this problem is to enhance the quality of the image. This method is called super-resolving, and 
the purpose is to make it easier for classifiers to recognize the person in the image. Major shortcoming has been that these generative models do not preserve identity features of the person. Solutions have been 
introduced to tackle this problem~\citep{face}.

\subsubsection{Face hallucination}

Impressive results considering face recognition have been obtained with SphereFace and ArcFace 
methods. The main flaw of these methods is that they perform poorly with low resolution images. 
Super-resolving an image from low-resolution (LR) to high-resolution (HR) is called super-resolution 
(SR). One example of SR tasks is face hallucination. Face hallucination methods can be used in face 
recognition with low resolution images. With face hallucination methods HR image can be 
reconstructed from LR image and the reconstructed image can be fed to a face recognition model~\citep{face}.

Facial images are usually different from other images since the face usually covers most of the 
image. Due to the occupations of the image, facial features should be taken into account while using face 
hallucination methods to improve the performance of the face recognition model. In recent years, 
identity-preserving methods have attracted attention. The main shortage of these methods has been 
that they perform poorly with other images of the same identity~\citep{face}.

Many methods have been proposed in the field of image super-resolution and most prominent 
methods are based on deep learning. Methods such as Very Deep Super Resolution Network (VDSR), 
Deeply-Recursive Convolutional Network (DRCN), Super-Resolution Generative Adversarial Networks 
(SRGAN) have been introduced with different approaches~\citep{face}.

A method called C-Face network based on SRGAN is proposed by \citet{face}. SRGAN includes Generative Adversarial Networks (GAN), Convolution Neural Network (CNN) and VGG19 network 
with three loss functions: A-softmax loss, Adversarial loss, and Perceptual loss. In addition to SRGAN, the 
C-Face loss function is introduced. The main purpose of the C-Face loss function is to ensure that the identity features of the person are preserved. 
Based on the results, the proposed solution outperforms other comparable solutions such as standard SRGAN and VDSR.

\subsection{3D point cloud classification and segmentation}

Beneficial data such as geometric and shape information can be obtained from 3D point clouds. 
This information can be used in various applications such as automatic driving. Due to the 
unstructured nature of the received data, exploiting the information is challenging. Different solutions have been
researched to process the data and make it more feasible for learning algorithms~\citep{3D}.

\subsubsection{Exploiting local attention features}

Convolutional Neural Networks (CNN) have been used with tasks related to 3D point clouds. Using CNNs usually 
leads to complicated calculations and information loss. Promising results have been 
obtained from PointNet++ network, which is an expansion of PointNet network. However, 
PointNet++ does not consider the spatial structure of the local area. To solve this problem different 
solutions are introduced; however, these solutions are prone to ignoring the importance of the local 
features, which dilutes the accuracy of 3D point cloud related tasks~\citep{3D}.

Several approaches have been introduced to solve 3D point cloud tasks such as multi-view\-point 
cloud feature learning, voxel‑based point cloud feature learning, learning features 
from unstructured point cloud directly, learning features based on graph theory, and learning 
features based on attention mechanism. Some of these approaches have shown promising results 
but many of them suffer shortcomings such as complex computation or loss of information~\citep{3D}.

\citet{3D} introduce Spatial Depth Attention (SDA) network was, which fuses global features with local attention features. Local attention features are 
features that considerably contribute to the final decision and are the most informative parts of the 
local feature scape. The spatial transform module was used to standardize the point cloud data. Two different feature extractor modules were 
used to extract the global features and the local attention features. Concat module was used to fuse the 
extracted features and this fused information was fed to the task execution module. Task execution 
module performs the desired task such as classification or segmentation. SDA was compared against solutions such as PointNet++ and LAM-Point++.
Based on the results, SDA outperforms other solutions in the classification task. In segmentation task SDA showed robust performance and was one of the 
best performers. 

\subsection{Event prediction}

Spatio-temporal event data contains time and location. This data shows when and where a certain 
event has appeared. For example, disease outbreaks are recorded as sequence of events based 
on location and time. Disease outbreak is a good example of an event that can cause secondary 
events which may spread to a wider epidemic. The disease outbreak has been caused by preceding 
triggering events, which have led to more extensive consequences. Considering disease control, it is 
crucial to health authorities to be able to predict when and where an outbreak will occur and what 
kind of events will trigger it. Event prediction has been highly researched and considering external features while processing spatio-temporal data have proven crucial to obtain
all the useful information~\citep{spatio}.

\subsubsection{Exploiting external features}

Hawken process, which is a general mathematic framework, has been used to model events such as 
infectious diseases and earthquakes. Models that use Hawken process do not consider external 
factors. These external factors, which impact the triggering process, include for example weather and 
population distribution. Today, a wide range of external information is available. For example, 
open-source platforms such as GIS provide useful information. In several studies Hawkes process has been extended 
to include external factors. However, the introduced methods usually require hand-crafted features 
that cannot utilize unstructured data. Unstructured data such as images may 
contain useful information which should be exploited~\citep{spatio}.

\citet{spatio} introduce a new architecture for Hawken processes called Convolutional Hawkes process 
(ConvHawkes). In ConvHawkes, the Hawkes process intensity is designed based on a convolutional 
Neural Network (CNN), which uses continuous kernel convolution. The proposed solution exploits
external features in georeferenced images to improve prediction performance. ConvHawkes has two components external effect and spatio-temporal decay. In 
external effect component CNN architecture transforms each image into a latent feature map. 
Continuous kernel convolution is used to expand the latent feature map onto the continuous spatio-temporal space. 
The purpose of the external effect component is to capture information from the 
external factors. With the external effect component, the neural network model is incorporated into 
formulation of Hawkes process. ConHawkes was compared against solutions such as Spatio-temporal homogeneous Poisson Process (SPP) and Recurrent Marked Temporal Point Process (RMTPP).
Based on the results, ConHawkes outperform other solutions and was the most consistent of them.

\section{Robotics and motion}

While developing autonomously operating systems, trajectory prediction and motion planning play a crucial role~\citep{trajectory},~\citep{CCTMP}. For example, trajectories of external factors such as pedestrians need to be predicted by self-driven cars to avoid
accidents~\citep{trajectory}. Via motion planning the self-driven car is prepared for different scenarios that can occur including variety of uncertain factors~\citep{CCTMP}. In this study, tasks and problems related to the motion planning and trajectory prediction of autonomous systems are included in this branch. The structure of the branch is illustrated in Fig 4. 

\begin{figure*}[t]
  \centering
  \includegraphics[width=7cm, height=6cm]{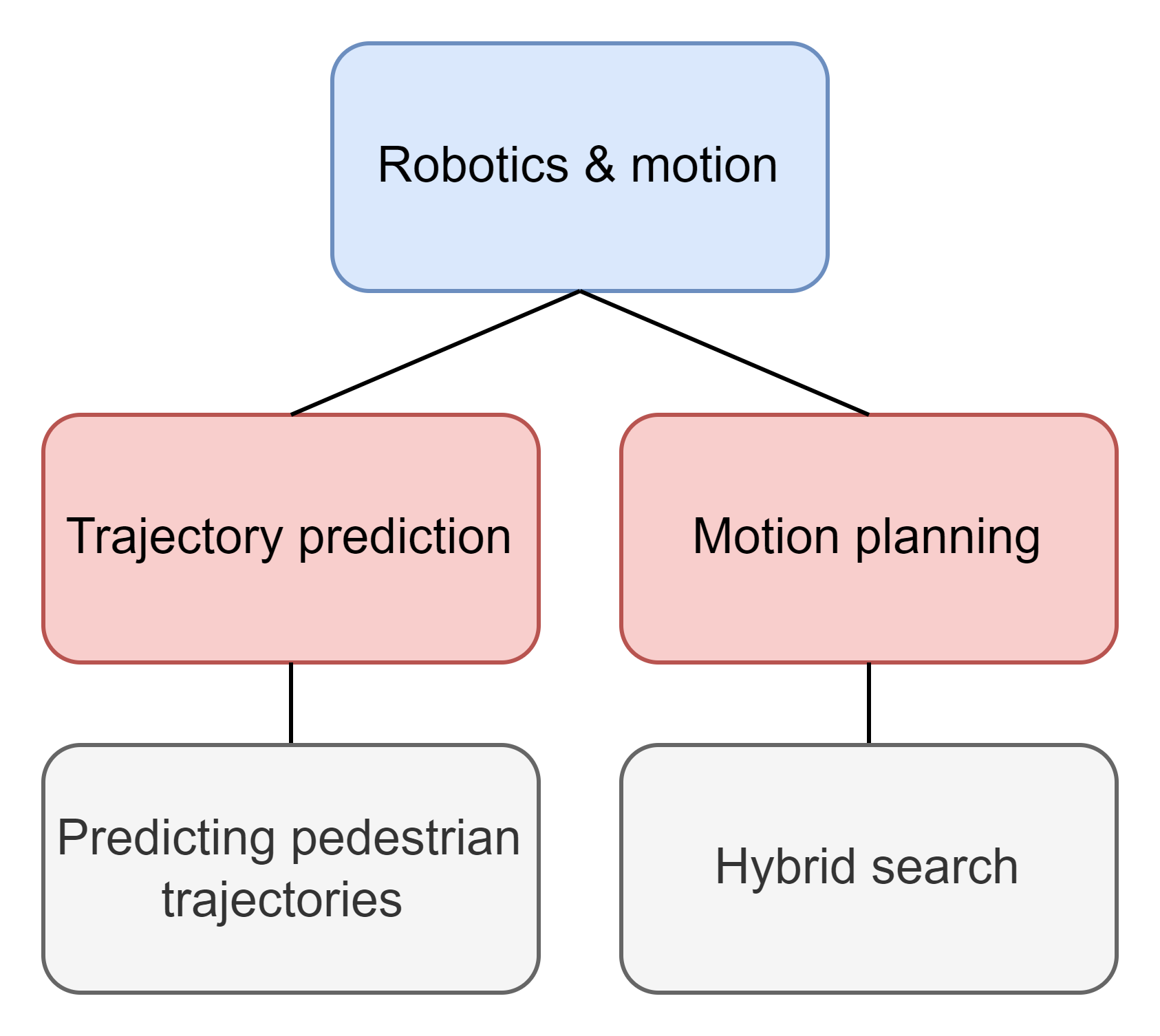}
  \caption{Structure of the Robotics and motion branch.}
  \label{Fig 4}
\end{figure*}

\subsection{Trajectory prediction}

Trajectory prediction is crucial while developing autonomous vehicles. Many studies have concentrated on pedestrian trajectories which autonomous cars must be able to predict. Changes in urban planning have
caused challenges for predicting these trajectories. Algorithms that can produce fast and accurate predictions are essential for adoption of autonomous cars~\citep{trajectory}.   

\subsubsection{Predicting pedestrian trajectories}

Pedestrian safety is a major concern while developing autonomous vehicles (AV). The common approach 
in development is to predict if a pedestrian is going to cross the road or not in front of the AV. A new 
way of planning urban areas is bringing vehicles and pedestrians into a shared space. Moving in a 
mutual space brings new challenges for AV development. When navigating in spaces where vehicles 
and pedestrians are mixed, AV must be able to predict pedestrian trajectories in different scenarios~\citep{trajectory}.

For simulating scenarios in a shared space, expert models and data-driven models have been 
developed. One example of an expert model is Social Force Model (SFM). SFM is widely used, but it is 
not meant for predicting or reproducing individual trajectories. Data-driven models provide more 
accurate results. A shortcoming of data-driven models is that they need plenty of data to become robust. 
Data from shared spaces is scarce, and these models do not scale for new situations. AVs need to 
produce real-time predictions and the predictions must be explainable. In crowded spaces a trajectory 
must be produced for each pedestrian. The trajectories must be accurate and produced fast~\citep{trajectory}.

An Agent-Based Modeling (ABM) approach is introduced by~\citet{trajectory}. The ABM can model how a pedestrian reacts to an AV in 
a shared space. An agent represents every individual pedestrian, and multiple agents are used to produce a crowd.
The pedestrian motion behavior includes three steps. In the first step, the agent observes its environment 
and perception is updated depending on what it perceives in its perception zone. The perception zone 
might include other pedestrians, static obstacles, or AVs. The perceptions are used to compute the future 
movement. SFM uses physical forces to compute this. The internal motivations of the 
pedestrian are represented by these physical forces. These physical forces include an internal 
attractive force, a repulsive force, a social force, and small random forces to avoid the agent to be too 
rigid. For example, a repulsive force is used to repel a pedestrian from going too close to an AV. These 
forces are used to perform actions. The agent tries to reach the destination without colliding with 
any obstacle. Situations where agents travel in a group, the agents try to stay together with other 
members. Four social groups are considered: couples, families, friends, and coworkers.
Time-to-conflict (TTC) is used to separate the interactions including a conflict with non-conflict 
ones. The distance is predicted, and based on the prediction it is decided if a conflict occurs. SFM is used 
for interactions with no conflict. The aforementioned repulsive force coming from an AV will repel 
a pedestrian from going too close. Decision model is used for conflict interactions. For example, 
decision model is used to decide if the agent should stop, run, turn or step back in a case of conflict. 
The decision made by the decision model modifies the forces used by SFM. The solution was compared against standard SFM.
Based on the results, the diversity of pedestrian behavior is better captured by ABM and the collision rates are more realistic. 
It was concluded that the proposed solution could be used in real-world applications.

\subsection{Motion planning}

Uncertainty is a major challenge for motion planning. Autonomous cars drive in uncertain 
surroundings while making quick decisions, and autonomous underwater vehicles face uncertain 
ocean currents. A vast amount of scenarios that could happen during the task must be handled by the 
agent. Motion planning has many predetermined requirements, and solutions such as hybrid search have been proposed to meet these requirements~\citep{CCTMP}.

\subsubsection{Hybrid search}

Due the large number of scenarios to be handled, motion planning can become 
computationally intensive. In case of autonomous underwater vehicles such as gliders, many 
requirements are set for motion planning for example a predetermined rate for mission failure, goals 
that the vehicle must accomplish, a period in which the solution should be produced, and optimizing of energy usage.
In recent years hybrid search has seen interest in motion planning. Hybrid search consists of more 
than one component and can be composed for example with a region planner and a trajectory 
planner. Hybrid search was proven effective in real world scenarios, such as underwater vehicles, by 
Hybrid Flow Graph which was introduced by Kongming. Although Kongming's solution does not 
consider the risk aspect and the focus of planning is simple linear systems that works in 2D 
environments. Other examples of hybrid search applications are The Scotty planner, pSulu planner~\citep{CCTMP}.

\citet{CCTMP} proposed a hybrid search solution with two components,
a region planner and a trajectory planner. The region planner is an upper-level planner, and the 
trajectory planner is a lower-level planner. The decision of which regions should be explored is made by the region planner. The selected region serves as a 
potential candidate for a trajectory. The trajectory can go between the original state and a goal state or 
between two goal states. The region planner gives an output containing a set of regions which 
are ordered, also called a path, to the trajectory planner. The trajectory planner consists of three 
models of risk: CDF-Based Chance Constraints, Sampling-Based Chance Constraints, and Shooting 
Method Monte Carlo (SMMC). The trajectory planner uses regions from the region planner to 
optimize trajectory which is constrained to go through these regions. The inputs given to the 
region planner at the beginning are an agent, goals that the agent must achieve, and a set of obstacles. 
The proposed solution was compared against solutions such as Chance-Constrained Rapidly-Exploring Random Tree (CC-RRT) and Disjunctive Linear Programming (DLP).
Based on the results, the proposed solution produced feasible solutions faster than comparable solutions.

\section{Reinforcement learning}

Reinforcement learning (RL) is a branch of artificial intelligence, where the artificial intelligent agent interacts with an unknown environment. While the agent interacts with the
environment, the state of the environment changes and the agent receives rewards based on the state. It is considered that RL agents are the most advanced artificial intelligent
agents~\citep{RL}. In this study, tasks and problems containing reinforcement learning agents are included in this branch. The structure of the branch is illustrated in Fig 5.

\begin{figure*}[t]
  \centering
  \includegraphics[width=\textwidth]{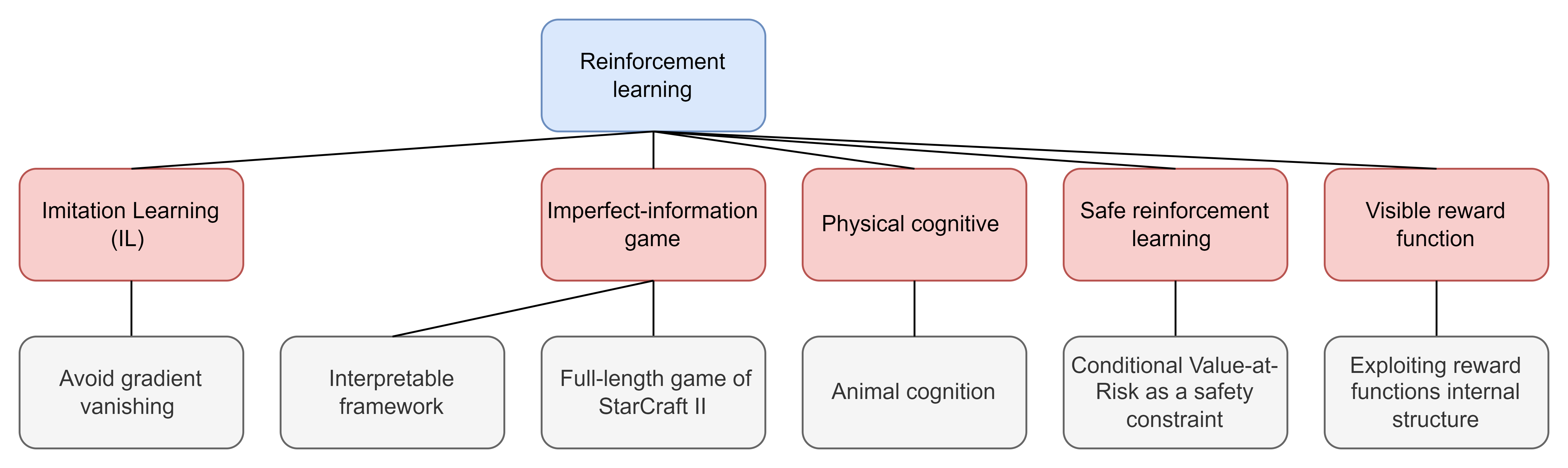}
  \caption{Structure of the Reinforcement learning branch.}
  \label{Fig 5}
\end{figure*}

\subsection{Imitation learning}

Imitation Learning (IL) has shown promising results while applied to domains such as robotics and 
Natural Language Processing (NLP). IL matches learning policies and actions of expert data. Acquired 
limited expert knowledge is used to recover policy. Inverse Reinforcement Learning (IRL) has been a 
promising approach for IL. IRL has shown some performance breakthroughs when generative-discriminative
framework is used. The imbalance between the generator and the discriminator has been 
a major problem with these frameworks. Different solutions have been introduced to overcome this issue~\citep{avoiding}.

\subsubsection{Avoiding gradient vanishing}

Imbalance between the generator and the discriminator occurs especially with Generative Adversarial 
Imitation Learning (GAIL) and is a severe problem while training, which is due to the different learning 
speed between the generator and the discriminator. The generator is a Reinforcement Learning (RL) agent that produces state-action pairs.
The discriminator applies supervised learning and learns much faster than the RL generator. This leads to a situation where a well-trained discriminator is obtained but the 
performance of the generator is poor~\citep{avoiding}.

\citet{avoiding} introduce an approach which tries to obtain a well 
trained generator with a discriminator, which acts more like a teacher. The discriminator is giving 
proper fake rewards to the generator to avoid gradient vanishing. The proposed solution is called 
Generative Adversarial Imitation Learning with Variance Regularization (GAIL-VR). The main issue 
with GAIL is that every state-action pair produced by the generator can be figured out by the discriminator due to the difference in learning speed. Leading 
to a situation where the generator gets only low rewards and variation is minor. To get higher rewards 
the generator should be able to produce state-action pairs, which the discriminator cannot distinguish from 
the expert data. There have been efforts to solve this problem by optimizing the discriminator algorithm to remove the gradient vanishing phenomenon. 
GAIL-VR was compared against solutions such as standard GAIL and Wasserstein Adversarial Imitation Learning (WAIL).
Based on the results, GAIL-VR obtains the best average reward and training speed almost with every MuJoCO environment.
It was concluded that GAIL-VR performs well with low and high dimensional state-action space. The proposed solution was also
very stable and same parameters can be used with different environments.

\subsection{Imperfect-information game}

Imperfect-information games such as StarCraft II and HUNL Poker are useful problems for reinforcement learning algorithms~\citep{starcraft},~\citep{interpretable}. 
A tremendous amount of state and action spaces provides challenges where a huge amount of decisions need to be made in a sort period of time~\citep{starcraft}. 
Different solutions have been researched, and progress has been made to solve these problems more efficiently~\citep{starcraft},~\citep{interpretable}. 
One major hindrance for reinforcement learning algorithms has been the lack of interpretability. The interpretability problem needs to be tackled before mainstream adoption~\citep{interpretable}.

\subsubsection{Full-length game of StarCraft II}

Deep reinforcement learning (DRL) and Deep Q-Network (DQN) have shown significant potential in 
solving problems such as the game of Go, Atari games, and self-driving vehicles. However, large-scale 
problems still appear to be challenging for RL algorithms. StarCraft has 
attracted attention as an environment for exploring the abilities of RL algorithms. As an environment 
StarCraft has a large map with huge state and action spaces. StarCraft provides imperfect 
information, and thousands of decisions need to be made in 10-30 minute periods. It is a multi-agent
game and cooperation with other players might be needed. The learning platform used with 
RL algorithms is called StarCraft II Learning Environment (SC2LE) which is based on StarCraft II (SC2)~\citep{starcraft}.

The most promising results concerning RL in game play of SC2 have been achieved by AlphaStar. Some 
arguments have appeared that the human knowledge used by AlphaStar is too dominating. AlphaStar 
does not create any new tactics and only uses tactics envisages by a human. Because of this and some 
other reasons, it has been argued that SC2 problem has not been perfectly solved by AlphaStar. Two 
open-source works can be found using an architecture similar to AlphaStar: DI-Star and SC2IL~\citep{starcraft}.

Hierarchical reinforcement learning (HRL) algorithm creates several sub-problems from a complex 
problem, and these sub-problems are solved one-by-one. The curse of dimensionality is a situation where 
state-space has a huge dimension and there is an exponential growth in explorable states. The curse of 
dimensionality can be solved by using HRL due to the dismantle of the problem in smaller sub-problems.
Some traditional HRL tools are Option, MaxQ, and ALisp. Some HRL algorithms that 
have been proposed in recent years are Option-Critic, FeUdalNetwork, and Meta Learning Shared 
Hierarchies (MLSH). MLSH is based on meta-learning, and it has defeated PPO algorithm in some 
tasks~\citep{starcraft}.

HRL solution was proposed by~\citet{starcraft} to solve the SC2 problem. The solution contains two different timescales with two types of policies. In long time interval, 
the controller gets a global observation and chooses a sub-policy based on that observation. In short 
time interval, the sub-policy chosen by the controller, selects a macro action. Every sub-policy has its own 
action-space and reward targets, and receives its own local observation. The architecture is a two-layer
architecture, but more layers can be inserted under the sub-policy layer for example to make it
a three-layer architecture. HRL makes it possible to split huge state and action spaces into smaller 
ones. When sub-policies have their own action spaces, training the algorithm is easier. Curriculum learning was also applied in the training phase. The curriculum was designed from easy to hard 
and used to help agents train on different difficulty levels. The controller was set to select a sub-policy
every 8 seconds, and macro-action was performed once in every second by the sub-policy. HRL algorithm was 
tested against non-hierarchical architecture where controller and sub-policies were switched off. 
The results indicate that HRL algorithm will perform significantly better than the non-hierarchical when 
the problem becomes more difficult. The proposed solution was compared against TStarBots and mini-AlphaStar (mAS), which consist of almost every component of AlphaStar. Based on the results,
TStarBots performed slightly better than HRL solution but required much more human knowledge and computing resources. It was concluded that
HRL solution is significantly more resource efficient and performs better with limited computing resources than comparable solutions.

\subsubsection{Interpretable framework}

Lack of interpretability is a huge drawback considering reinforcement learning (RL) agents. This is a 
major concern and if not solved, it will prevent mainstream adoption of the technology. Use of neural 
networks is considered one of the reasons RL agents become hard to interpret~\citep{interpretable}.

Interpretable RL-agent is researched by~\citet{interpretable}. 
The study compares three different learning algorithms Optimal Classification Trees (OCT), Extreme 
Gradient Boosted Trees (XGBoost), and Feedforward Neural Network against each other and against 
Slumbot in the game of HUNL Poker. A novel approach is introduced in transforming game state to a 
vector, which is a combination of two components. Counterfactual Regret Minimization (CFR) self-play algorithm is used to find the average strategy in 
HUNL Poker. The aforementioned three algorithms try to learn this strategy found by CFR. As a presumption, OCT can be denominated as the most interpretable and 
Feedforward Neural Network as the least interpretable. XGBoost can be considered as non-interpretable
but not as much as Feedforward Neural Network. The study tried to create a framework where the whole process from generating features to model 
training is as interpretable as possible. Based on the results every algorithm outperforms Slumbot. Feedforward Neural Network based agent 
is the best performer closely followed by XGBoost. OCT based agent is not far behind from Neural Network based agent and uses far less parameters.
It was concluded that it is possible to create a very powerful HUNL Poker agent with an interpretable framework. 
Using OCT as a learning algorithm, human-readable prints can be produced to analyze the strategy the algorithm uses.

\subsection{Physical cognitive reasoning}

Deep reinforcement learning (DRL) systems have shown promising results when used in complex 
games. However, the ability to imitate basic human cognitive skills such as spatial reasoning is usually 
lacking in these systems. Animal-AI (AAI) has been used as a testbed in artificial intelligence 
competitions when evaluating physical cognitive reasoning~\citep{physical}.

\subsubsection{Animal cognition}

Results have shown that the top performing systems utilizing DRL failed to solve 
tasks such as spatial elimination, which requires physical reasoning with common sense. DRL systems are 
opaque, and they generalize poorly with unseen samples. These flaws are inherited from neural 
networks, which are used in DRL systems as function approximators. A great number of studies have been conducted considering merging neural systems with symbolic systems. 
Promising results have been obtained as symbolic systems usually increase the performance of DRL 
methods and provide better interpretability~\citep{physical}.

The Detect, Understand, Act (DUA) approach was introduced by~\citet{physical}. Two different levels of operations were used in the 
DUA: micro-level and macro-level. Micro-level is for temporal abstraction and mapping 
environmental observation for discrete actions. Macro-level is a timescale which consists of a huge 
amount of environment timesteps, and it maps symbolic states to options. The proposed solution is composed of three components: Detect, Understand, and Act. At each 
timestep, the Detect module filters information received from the environment to a meaningful 
representation. This meaningful representation also known as symbolic representation is processed 
by the Understand module. The Understand module uses the learned meta-policy to initiate the 
correct option. These options are pre-trained DRL agents, and they are included in The Act module. 
The Understand module gives instructions on how to filter the input information that is fed to these 
DRL agents. The Detect module filters raw images into more useful features. Images are parsed into a set of 
bounding boxes. Information from these bounding boxes is translated into an Answer Set 
Programming (ASP) program. The Understand module contains two sub-modules: ASP program, and 
Inductive Logic Answer Set Programming (ILASP) learner. Inductive Meta-Policy learning (IMP) is used to learn the symbolic meta-policies used in the Understand module. 
The Act module uses Proximal Policy Optimization (PPO) as DRL algorithm. DUA was compared against the best performers in 2019 AAI competition. Based on the results the proposed solution
provided state-of-the-art performance and would have become third in 2019 AAI competition. It was concluded 
that DUA is able to provide interpretable results and it can be generalized for more complex tasks.

\subsection{Safe reinforcement learning}

While training Reinforcement Learning (RL) agent in simulated environments, a huge amount of 
interactions with the environment is required. This can be a major burden when trying to train physical 
systems in real-world environments. In real-world environments safety issues must also be 
considered more thoroughly and it might be feasible to integrate safety aspect into the systems 
behavior. For example, autonomous drone trajectories can be constrained while flying in an 
environment that contains people. The real-world settings require a framework, which includes 
sample-efficiency and safety aspect. There are two different approaches considering safety. It can be 
required that the controlling agent/policy is safe or that the learning process is safe~\citep{safe}.

\subsubsection{Conditional Value-at-Risk as a safety constraint}

A solution called Safe Model‑Based and Active reinforcement learning (SAMBA) is introduced by~\citet{safe}. 
SAMBA uses Conditional Value-at-Risk (CVaR) as a safety constraint and is appropriate to use in 
continuous state and action space for learning control. In the structure of SAMBA the PILCO is used as a 
template which is altered with two novel methods allowing active exploration and safety. SAMBAs 
focus is to minimize cost, maximize active exploration, and conform safety constraints. Samba was compared against unconstrained model-free
algorithms such as PPO, expectation constrained model-fee algorithms such as Safety-constrained PPO (SPPO), and unconstrained model-based algorithms such as PlaNet.
Based on the results, SAMBA reduces samples and Total cost (TC) used in training. SAMBAs safety performance was as good or better when compared to other solutions.
It was also concluded that sample efficacy does not undermine the safety during the testing phase.

\subsection{Visible reward function}

Reinforcement Learning (RL) agent has to learn from its experience by interacting with the environment. 
A common assumption is that RL agent does not know its reward function. This means that the agent 
only makes queries to the function based on the current situation, although the reward function 
may have some high-level ideas which could be utilized by the agent. Access to the structure of the 
reward function could provide a better and faster learning result~\citep{exploiting}.

\subsubsection{Exploiting reward functions internal structure}

An approach called reward machine is 
introduced by~\citet{exploiting}. The reward machine is composed of different reward functions. When an agent transits from 
state to state inside an environment, it does the same thing within the reward machine. The reward 
function that the agent should use is the output by reward machine based on the state. The agent 
knows the amount of states available and can use this information while learning. The reward 
machine was applied to two structures: Counterfactual experiences for Reward Machines (CRM), and 
Hierarchical RL for Reward Machines (HRM). CRM uses experiences it has already learnt to achieve the right behavior at different reward machine 
states. For example in a task where the agent should pick up a coffee before it reaches the destination; if the agent reaches the destination before it picks 
up the coffee it will not get the reward. CRM will use this knowledge of how to reach the 
destination as soon as it finds the coffee. With HRM the problem is decomposed into subproblems 
called options. For example, if the task is to pick up coffee and mail 
before reaching the destination, HRM would learn several policies such as picking up the coffee 
before the mail and picking up the mail before the coffee. Additionally, a method called automated Reward Shaping (RS) is introduced the core idea of which is to provide intermediate rewards for the agent while 
learning the task. RS is applied with CRM and HRM. Q-learning (QL), Double Deep Q-Network (DDQN), and Deep Deterministic Policy Gradient (DDPG) were used as a core off-policy learning methods
in different test scenarios. CRM and HRM were compared against each other and against vanilla versions of core off-policy learners. Based on the results, in most of the test environments CRM
outperforms other solutions. In an environment where the action and the state space were continuous, HRM was the best performer. Proposed RS improved the performance in some test environments. 
It was concluded that although the proposed solutions outperformed the baseline solutions in every experiment, CRM and HRM were computationally more expensive.

\section{Others}

This section contains other artificial intelligence technology solutions and tasks that cannot be distinctly inserted in the aforementioned branches, including solutions such as solvers that
solve the satisfiability problem (SAT)~\citep{sat} and solutions that predict survival outcomes~\citep{survival},~\citep{deep_survival}. However, these solutions may include some aspects of the previous branches.
The structure of the branch is illustrated in Fig 6.

\begin{figure*}[t]
  \centering
  \includegraphics[width=\textwidth]{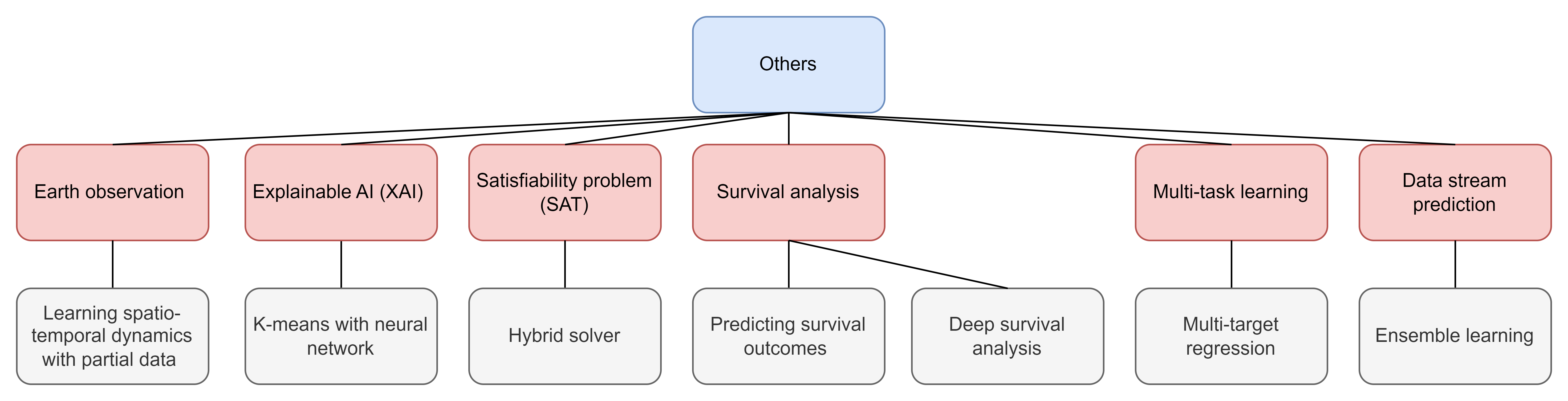}
  \caption{Structure of the Others branch.}
  \label{Fig 6}
\end{figure*}

\subsection{Earth observation}

Mostly in earth science, machine learning solutions have been used to solve a specific problem. The 
progress in deep learning has provided better integration of these two domains, although the earth 
science has some specificities, which makes it very challenging for deep learning solutions. The earth 
science includes complex time-evolving multidimensional structures, and the observations are often 
at different spatio-temporal resolutions. Usually, the observations are raw data and unlabeled. In 
most cases the full state of the system is not observable and partial information is available; for 
example when observing oceans with satellite images, subsurface information is difficult to observe. 
Generalization is usually not possible since the earth science problems are usually underconstrained~\citep{earth}.

\subsubsection{Learning spatio-temporal dynamics with partial data} 

A new framework is introduced by~\citet{earth} for learning spatio-temporal dynamics. The framework operates with partially observable data and in 
large observation spaces. Two different settings are introduced, one which only relies on 
observations and one which assumes that each trajectory has full initial state. The solution uses Convolutional Neural Networks (CNN) to learn spatial 
differential operators from data without supervision. CNN parametrizes the forward equation to obtain the time evolution. The proposed framework was compared against 
PKnI and PRNN. Based on the results, the framework outperformed other solutions. The setting that assumed that each trajectory has full initial state performed slightly better
than the one, which only relies on observations.

\subsection{Explainable AI}

Clustering is a method where similar samples are grouped into the same cluster. Recent focus, 
considering clustering, is towards high-dimensional data handling. Many solutions have been 
introduced such as subspace clustering and deep clustering. Although providing promising results, 
these solutions suffer from shortcomings such as limited representability and lack of explainability~\citep{explainable}.

\subsubsection{K-means with neural network}

A novel solution is introduced by~\citet{explainable} where 
vanilla k-means is equipped with a neural network. The proposed solution is called inTerpretable 
nEuraL cLustering (TELL). With TELL, an interpretable neural network is built for unsupervised 
learning tasks, more specifically, clustering tasks. In TELL's structure, k-means discrete objective is 
reformulated as a neural layer. This structure makes it more applicable to large-scale and online 
data. In TELL the cluster layer is explainable including the input, weights, activation, and loss function. Also, it is possible to reason TELL's error surface and 
dynamic behavior mathematically. TELL was compared against solutions such as subspace clustering solutions Spectral Clustering (SC) and large-scale clustering solutions Scalable LRR (SLRR).
Based on the results TELL was significantly better than other solutions in most of the test cases. Two visualization analyses were conducted, and it was concluded that as the training goes, TELL 
improves the performance and more discriminative representations are learned. 

\subsection{Satisfiability problem}

The satisfiability problem (SAT) solvers have been used to solve problems from different domains 
such as electronic design automation (EDA), hardware verification, and mathematical theorem 
proving. SAT solver tries to solve satisfiability of a given propositional formula, which is often in 
Conjunctive Normal Form (CNF). The most popular SAT solvers are Conﬂict Driven Clause Learning (CDCL) 
based solvers and local search solvers. CDCL based SAT solvers have shown significant success in 
many practical applications and local search solvers have shown good performance in random and 
hard combinatorial cases. Satisfiability modulo theory (SMT) also uses SAT solvers as a core 
component. Promising results as solving SAT have been gained from CDCL solvers such as Glucose, MapleLCMDistChronoBT-DL, Kissat and CaDiCaL. Several gold medals have been won by Glucose in SAT Competitions and it is considered as a 
milestone in CDCL solver development. SAT Race 2019 was won by MapleLCMDistChronoBT-DL  
and Main Track of SAT Competition 2020 was won by Kissat. Considering local search solvers CCAnr 
has shown good performance on structured instances~\citep{sat}.

\subsubsection{Hybrid solver}

Shortcomings in CDCL solvers, such as frequent restarts, have been tried to be solved by combining a CDCL 
solver and local search solver. Often these hybrid solvers fail in communication, and only partial 
information is sent to one direction. Therefore, these two solvers consider each other as a black box 
and the hybrid solver does not perform effectively~\citep{sat}.

\citet{sat} created a hybrid which includes a CDCL solver and a local search solver. The main solver is the CDCL solver, and the local search 
solver is used to enhance branching heuristics in the CDCL. A frequent information flow is 
established between the two solvers and the information moves in both directions. Several hybrid solvers were created and Glucose, MapleLCMDistChronoBT-DL, Kissat, and CaDiCaL were used as a CDCL solver. CCAnr was used as a local search 
solver. Their performance was compared using the main track benchmarks of the SAT Competitions from 
2019 to 2022 and SAT Race 2019. Based on the results the hybrid solver considerably improved the performance of the CDCL solvers. 

\subsection{Survival analysis}

To predict the expected duration for some event to occur, a survival analysis is performed. Compared 
to other analytical methods survival analysis is more challenging because of censoring. Censoring 
means that the actual event time is different than the observed time because the actual event has not been observed. In case of censoring, it can only 
be known that the actual time is greater or lesser than the observed time~\citep{survival}.
Real-world datasets used in survival analysis often contains censored data.
Lack of observation might be caused by situations such as missing follow-up. In these cases, a censored time is recorded which can be for example the time 
when the observation window has ended. This gives the information that an event has not occurred 
before the recorded time. These recordings are utilized by survival analysis methods~\citep{deep_survival}. 

\subsubsection{Predicting survival outcomes}

In survival applications random survival forests (RSF) are widely used, and they have shown 
superiority compared to traditional survival models. In separation to original random forest, RSF 
considers censoring information and survival time explicitly. Semi-supervised learning (SSL) methods, which take advantage of the unlabeled data, are also 
used in survival analysis, but their full potential is yet to be discovered~\citep{survival}.

Self-trained random survival forest (ST-RSF) is a semi-supervised learning algorithm, which exploits 
self-training wrapper technique. The dataset is divided into three sets: observed, censored, and unlabeled, 
where the observed and censored are labeled sets. ST-RSF uses labeled data to build an initial model. 
After the model has been built, ST-RSF iteratively increases the labeled set by choosing the most confident 
predictions in unlabeled set. The amount of unlabeled data to be added to labeled set is manually 
regulated. Iterative procedure is terminated by defining a global stopping criterion. Self-trained random survival forest corrected with censored times (ST-RSF+CCT) is similar to the 
ST-RSF algorithm except it exploits contained information from censored data. Censored 
data is treated as unlabeled data in spite of the fact that censored data is not totally unlabeled. The information 
captured from censored instances is used to determine which unlabeled instances, including 
censored instances, will be added to the labeled data~\citep{survival}.

\citet{survival} conducted a research, where ST-RFS and ST-RFS+CCT were compared against each other and against Standard random survival forest (RSF) and Cox regression with LASSO regularization 
(Lasso-Cox). Based on the results, both ST-RFS and ST-RFS+CCT, outperform standard RSF and Lasso-Cox. ST-RFS+CCT also manages to outperform ST-RFS and showed more steady behavior.

\subsubsection{Deep survival analysis}

Due to the ability to extract features from a vast amount of raw data, deep neural network solutions 
have been researched considering survival analysis. Likelihood-based methods are commonly used 
options for estimating a probabilistic model. These methods are also extensively used for deep 
survival analysis; however, censored data inflicts challenges while using neural network models due 
to the difficult-to-evaluate integrals. Most of the existing research has tried to avoid the integrals in 
different ways, but the introduced solutions have caused shortcomings such as limitations for flexibility 
and information loss~\citep{deep_survival}.

A solution called Survival model through Ordinary 
Diﬀerential Equation Networks (SODEN) is introduced by~\citet{deep_survival}. The SODEN uses Ordinary Diﬀerential 
Equation (ODE) and exploits ODE solver. SODEN is non-linear and makes no Proportional Hazard (PH) assumption, it is a continuous-time model, and 
Stochastic Gradient Descent (SGD)-based algorithms can be applied with it. SODEN was compared against solutions such as DeepSurv, and DeepHit.
Based on the results SODEN outperformed continuous-time based models such as DeepSurv. SODEN also achieved almost equal 
performance with the best performer DeepHit considering C-index, which measures the discriminative performance. Considering IBS and IBLL, which measure the combination of the 
calibration performance and the discriminative performance, SODEN showed the best performance almost with every dataset.

\subsection{Multi-task learning}

Usually conventional machine learning (ML) algorithms predicts only a one target. However, there 
are several tasks that demand multi-target predictions. Each target can be provided with its own 
independent ML algorithm. When using independent algorithms, the information cannot be shared 
between the algorithms. Multi-task learning (MTL) algorithms are used for implementing multiple 
relevant tasks. The tasks are implemented simultaneously, and multiple outputs are predicted at the 
same time. Due to the relevance of the tasks, information between the tasks is shared. With shared 
information the quality of predictions can be improved~\citep{MTL}.

\subsubsection{Multi-target regression}

Multi-target regression algorithms are a subset for MTL. Considering input samples, multi-target 
regression algorithms differ from MTL. In MTL, each task can have different samples and the number 
of samples can differ. Multi-target regression algorithms use same samples with each task. MTL 
algorithms can be used to solve multi-target regression problems but not the other way around. 
One shortcoming for MTL algorithms when solving multi-target regression problems is that they might 
require a complex structure. A more complex structure raises the calculation costs and might expose 
to overfitting. Using multi-target regression algorithms while solving multi-target regression 
problems can be more efficient due to the difference in input samples. Using the same input samples 
with each task can reduce the complexity of the algorithm~\citep{MTL}.

\citet{MTL} introduce a multi-target regression algorithm called Joint Gaussian Process Regression (JGPR). 
Most of the previous GPR algorithms are contributed to multi-task problems, which is why they are 
highly complex. The algorithm proposed in the study had lower complexity. Overfitting was prevented by using a shared covariance matrix. A shared covariance matrix provides the same hyperparameters for 
all the targets. JPGR was compared against solutions such as Multi-layer Multi-target Regression (MMR), and Multi-Object Random Forests (MORF).
Based on the results JGPR outperforms other solutions almost with every dataset. It was concluded that JGPR has a shortcoming with categorial variables.

\subsection{Data stream prediction}

Data streams is a notion that combines the volume and the velocity of the data. These streams are 
constantly evolving, which causes changes in their characteristics and definitions. This change is called 
concept drift, and it demands classifiers to update and adapt continuously. Combining a concept drift 
with class imbalance causes a major challenge for classifiers to learn. This combination exposes 
imbalance ratio and class roles to change dynamically. This challenge has been studied intensively 
and most prominent results have been achieved with ensemble learning~\citep{data_stream}.

\subsubsection{Ensemble learning}

There are three main approaches to apply while using ensemble learning to data streams: dynamic 
combiners, dynamic ensemble setup, and dynamic ensemble updating~\citep{data_stream}. \citet{data_stream} introduced a hybrid solution which 
combines these approaches, and this solution is called Robust Online Self-Adjusting Ensemble (ROSE). ROSE has four main features: variable size random feature 
subspaces, detection of concept drift and background ensemble, sliding window per class, and self-adjusting lambda for bagging. The ensemble consists of classifiers, and each base classifier is built on a 
random feature subspace. Using diverse feature subspaces with random size has proven 
effective in previous studies. ADWIN drift detector is used to detect concept drifts. If a concept drift is 
detected ROSE starts to train another ensemble. Another ensemble is trained in the background and 
it is not effected by old concepts. The classifiers compete with each other, and the worst performing 
classifiers will be discarded. When a new background ensemble is created a balanced class 
distribution is provided by sliding window per class. The sliding window per class makes the classifier 
more unbiased, and it will not favor the majority classes. The value of lambda is self-adjusted based on 
imbalance ratios. Self-adjustment enables ROSE to use minority class instances more effectively to train classifiers. 
ROSE was compared against class-imbalance ensembles such as OOB and OUOB, and general purpose ensembles such as KUE and SRP. HoeffdingTree was used as a base learner.
Based on the results, ROSE outperforms other solutions in many test cases. It was concluded that ROSE could effectively handle borderline and rare instances, remove noisy features very efficiently, and it
was highly flexible and stable.

\section{Discussion}


According to our study, the following branches have been identified as the latest trends in AI research. They were published in leading AI research journals in 2022, and as such represent the latest trends with real-world applications. The included articles were clustered into thematic branches, which contain similar domains and related topics. The branches are:

\begin{itemize}
\item Natural Language Processing
\item Computer Vision
\item Robotics and Motion
\item Reinforcement Learning
\item Others
\end{itemize}

Below, a summarizing overview of each branch is provided. More comprehensive explanations are provided in the previous sections. 

\subsection{Natural Language Processing}

Text classification is an increasingly relevant task, as vast text datasets are available. Detecting malicious content has real-world implications, as these threats have become everyday reality. At the same time, productivity improvements with text technologies are relevant. Tasks such as text summarization and text translation boost productivity in all domains. Furhtermore, sentiment analysis with modern technologies provides more accurate predictions for commerce and public services. In the future, large language models (LLMs) will assume some of these roles. 

Textual data is often imbalanced, and there is a lack of labeled datasets. Due to this, unsupervised learning and semi-supervised learning algorithms are often researched in relation to text classification.
Promising results have been obtained by using One-Class Learning (OCL) and Positive and Unlabeled Learning (PUL) algorithms. In tasks where a balanced and labeled dataset is available, supervised learning algorithms are preferred because these solutions usually 
perform better than unsupervised learning and semi-supervised learning algorithms. Pre-trained language models such as Bidirectional Encoder Representations from Transformers (BERT) and feature extractors such as bag-of-words are frequently used with supervised learning classifiers.
The best results considering supervised learning classifiers have been obtained by using a neural network and random forest. 
Multi-document Event Summarization (MES) frameworks are used in text summarization tasks. These frameworks can summarize the core event in few words.

In text translation tasks, Machine Translation (MT) frameworks are used to translate complete sentences into another
language. In real-time translation neural MT (NMT) frameworks are used. MT frameworks are also extended with multimodality which enables the exploitation of additional context such as images. Contextual information is also leveraged with NMTs as an additional input. 
This contextual information is processed using pre-trained language models such as BERT. In sentiment analysis, Aspect-Based Sentiment Analysis (ABSA) frameworks are highly used. ABSA is a multimodal structure where Long Short-Term Memory (LSTM) networks, 
Graph Convolutional Networks (GCN), and pre-trained language model BERT are used as context encoders. Improved performance has been obtained by adding an aspect aware mechanism to the context encoder.  

\subsection{Computer Vision}

Image recognition is a widely used technology in industrial processes and security products. Cyber security threats, such as deceitful artificial images, are also a problem, which need detection technologies. Nowadays, image generation is widely used in creative industries. Such realistic or stylistic generated images have acted as a productivity boost. Classification and segmentation are also widely used in industrial contexts. 

Combining language with image has often been used in image classification tasks. Significant results have been achieved by using Contrastive Language-Image Pre-Training (CLIP). Improved versions of CLIP have also been introduced, which enables updating CLIP without retraining the whole model.
In separating artificially generated images from real images, two architectures are commonly used; one that utilizes frequency space information and one that utilizes raw data. Improved performance has been obtained by fusing these architectures with an architecture that 
exploits wavelet packets. 

Considering face recognition, super-solving solutions have been used to enhance the quality of the image. Deep learning based solutions such as Super-Resolution Generative Adversarial Networks (SRGAN) and Very Deep Super Resolution Networks (VDSR) are highly used in this task. Enhanced quality makes it easier to a classifier to recognize the person in an image. 
Identity-preserving methods have also been used to increase the performance. In 3D point cloud classification and segmentation tasks PointNet++ is commonly used. With networks that fuse global features with local attention features such as Spatial Depth Attention (SDA), improved performance has been obtained.     

\subsection{Robotics and Motion}

Autonomous vehicles have been actively developed during the recent years. To make such vehicles safe for traffic, they must make accurate predictions. In addition, they must quickly decide optimal routing and motions. A safe and working autonomous vehicle would revolutionize city traffic. 

Simulating scenarios in shared space where vehicles and people encounter, expert models and data-driven models are commonly used. Utilizing expert model Social Force Model (SFM), solutions have been developed such as Agent-Based Modeling (ABM) that provides improved performance against vanilla SFM when predicting pedestrian trajectories.
Considering motion planning, hybrid search solutions have proven effective. Good results have been obtained by creating an architecture with two components, a region planner and a trajectory planner.

\subsection{Reinforcement Learning}

Software and robotic agents have become more feasible. Reinforcement learning is a way to train such agents using simulated environments. Various games, but also self-driving vehicles, are current application areas. However, large-scale problems are identified as challenges with these technologies. Perhaps the real-world usage is still to come. 

Inverse Reinforcement Learning (IRL) is a frequently used reinforcement learning (RL) approach. Performance breakthroughs have been achieved by using generative-discriminative frameworks such as Generative Adversarial Imitation Learning (GAIL). 
GAIL has also been improved by Variance Regularization (GAIL-VR), which enhances the imbalance between the generator and the discriminator. 
Considering imperfect-information games StarCraft II (SC2) has attracted attention as an environment. AlphaStar has shown the most impressive results while playing SC2. With a solution that exploits the Hierarchical Reinforcement Learning (HRL) structure, significant results have also been obtained. An interpretable framework
for imperfect-information game task has been achieved by Optimal Classification Trees (OCT) while playing HUNL Poker. 

Considering physical cognitive reasoning approaches, promising results have been obtained by merging neural systems with symbolic systems. Symbolic systems increase the performance of Deep Reinforcement Learning (DRL) methods and provide better interpretability.
Animal-AI has been used as a testbed in artificial intelligence competitions while evaluating physical cognitive reasoning solutions. Considering safe reinforcement learning, real-world settings require a framework which includes sample-efficiency and safety aspect.
Using Conditional Value-at-Risk (CVaR) as a safety constraint has shown promising results when applied in a practical, data-efficient model-based policy search method (PILCO) framework. Letting the RL agent utilize the structure of the reward function rather than just making queries has shown improvements in the overall performance of the RL agents. 

\subsection{Others}

Considering earth observation, promising results have been obtained with a framework that works with partially observable data. For explainable AI, a solution where k-means is equipped with a neural network has outperformed other comparable solutions. Considering the satisfiability problem (SAT), 
combining Conflict Driven Clause Learning (CDCL) solvers with local search solvers, performance improvement has been obtained. In survival analysis, RSF has shown superiority while predicting survival outcomes. Random Survival Forests (RSF) has also been extended with semi-supervised learning methods, and an improved performance has been
achieved. Considering multi-task learning, Multi-Task Learning (MTL) frameworks are widely used, although using multi-target regression framework for such problems has proven to be more efficient. Data streams demand classifiers to update and adapt continuously. Combined with class imbalance,
a major hindrance for classifier has been imposed. Most prominent results to tackle it have been obtained with ensemble learning.

\section{Conclusion}

Modern trends in artificial intelligence technology seem to respond to needs that arise from practical applications, including natural language processing, computer vision, robotics and agent-based learning. There are several distinctive points of interest in the current technology development in the field:

\begin{itemize}
\item There is a demand for solutions that can operate with sparse and imbalanced data from various different sources.
\item Solutions must be able to process more complex unstructured data, and manual work needs to be minimized. 
\item Learning should be continuous, and classifiers should be able to update without retraining the whole algorithm. 
\item Due to the demand for labeled data, unsupervised learning solutions and semi-supervised learning solutions are preferred over supervised learning solutions. 
\item Hierarchical and hybrid solutions are widely researched and different methods have been merged such as neural network systems and symbolic systems. 
\end{itemize}

Furthermore, it seems that the old areas of improvement are still in focus. 
(i) Obviously, there is a high demand for solutions that are computationally cost efficient and less complex. 
(ii) While operating in real-world environment, fast and accurate predictions need to be done. 
(iii) The safety aspect should be integrated into real-world systems and predictions should be explainable. 
(iv) Overall focus seems to be on creating scalable and generalizable solutions with interpretable predictions. 

The purpose of this study was to provide an overall view of the technology development and 
introduce various solutions for solving various real-world problems. 
This holistic view of the field of AI research is the main contribution of this article.
Researchers from various areas, such as health, manufacturing and energy sector, could benefit from the results, as they present possible methods and tasks that could be studied in detail within the area.
Limitations of this study concern the initial number of articles and the scope of journal selection, because there are various other journals in the field, some of them even having a high ranking in national lists. Therefore, many presented tasks and technological solutions only contain a single article. Furthermore, not every topic where AI is used is covered in the study.
Future work includes more concentrated research to better understand specific technologies. 
Artificial intelligence technologies are evolving rapidly, and more technological solutions are being developed to solve demanding tasks. 
As the ethical and societal issues related to AI are more recognized than before, the impact of new technologies should be studied. 
Due to the speed of development, this study could be extended in the future to provide a more extensive and up-to-date view. 




\section*{Authors' Contributions}

Teemu Niskanen, Tuomo Sipola, and Olli Väänänen contributed in conceptualization, methodology, and writing—original draft preparation. 
Teemu Niskanen contributed in investigation and visualization. 
Tuomo Sipola, and Olli Väänänen contributed in writing—review and editing, and supervision. 



\section*{Acknowledgments}

The authors would like to thank Tuula Kotikoski for proofreading the manuscript. 

\bibliography{recent_trends_in_ai_technology}


\section{Appendix}

Table~\ref{tab:abbr} lists all abbreviations found in this paper.

\begin{center}
\begin{longtable}{l l}
    \caption{List of abbreviations}
    \label{tab:abbr}\\
    \toprule
    Abbreviation & Meaning\\
    \midrule
    AA & Aspect Aware\\
    AAI & Animal Artificial Intelligence\\
    ABM & Agent-Based Modeling\\
    ABSA & Aspect-based Sentiment Analysis\\
    ACSA & Aspec Category Sentiment Analysis\\
    ADWIN & ADaptive WINdowing\\
    ASP & Answer Set Programming\\
    ATSA & Aspect Term Sentiment Analysis\\
    AI & Artificial Intelligence\\
    AV & Autonomous Vehicle\\
    BERT & Bidirectional Encoder Representations from Transformers\\
    BLEU & BiLingual Evaluation Understudy\\
    CCAnr & Configuration Checking with Aspiration non-random\\
    CC-RRT & Chance-Constrained Rapidly-Exploring Random Tree\\
    CDCL & Conflict Driven Clause Learning\\
    CDF & Cumulative Distribution Function\\
    CFR & Counterfactual Regret Minimization\\
    CITR & Control and Intelligent Transportation Research\\
    CLIP & Contrastive Language-Image Pre-Training\\
    CNF & Conjunctive Normal Form\\
    CNN & Convolutional Neural Network\\
    CoLLIE & Continual Learning of Language Grounding from Language-Image Embeddings\\
    CRM & Counterfactual experiences for Revard Machines\\
    CVaR & Conditional Value-at-Risk\\
    DCT & Discrete Cosine Transform\\
    DDPG & Deep Deterministic Policy Gradient\\
    DDQN & Double Deep Q-Network\\
    DLP & Disjunctive Linear Programming\\
    DQN & Deep Q-Network\\
    DRCN & Deeply-Recursive Convolutional Network\\
    DRL & Deep Reinforcement Learning\\
    DUA & Detect, Understand, Act\\
    DUT & Dalian University of Tecnology\\
    EDA & Electronic Design Automation\\
    FFHQ & Flickr Faces High Quality\\
    GAIL & Generative Arversarial Imitiation Learning\\
    GAIL-VR & Generative Arversarial Imitiation Learning with Variance Regularization\\
    GAN & Generative Adversarial Network\\
    GCN & Graph Convolutional Network\\
    GIS & Geographic Information System\\
    GPR & Gaussian Process Regression\\
    HR & High-Resolution\\
    HRL & Hierarchical Reinforcement Learning\\
    HRM & Hierarchical Reinforcement Learning for Reward Machines\\
    HUNL & Heads-Up No-Limit (poker game)\\
    IBS & Integrated Brier Score\\
    IBLL & Integrated Binomial Log-Likelihood\\
    IL & Imitation Learning\\
    ILASP & Inductive Logic Answer Set Programming\\
    IMP & Inductive Meta-Policy Learning\\
    IRL & Inverse Reinforcement Learning\\
    JGPR & Joint Gaussian Process Regression\\
    k-NND & k-Nearest Neighbors Density\\
    KUE & Kappa Updating Ensemble\\
    LASSO & Least Absolute Shrinkage and Selection Operator\\
    LR & Low-Resolution\\
    LSTM & Long Short-Term Memory\\
    MES & Multi-document Event Summarization\\
    ML & Machine Learning\\
    MLSH & Meta Learning Shared Hierarchies\\
    MMR & Multi-layer Multi-target Regression\\
    MMT & Multimodal Machine Translation\\
    MORF & Multi-Object Random Forest\\
    MRI & Magnetic Resonance Imaging\\
    MT & Machine Translation\\
    MTL & Multi-Task Learning\\
    NLP & Natural Language Processing\\
    NMT & Neural Machine Translation\\
    OCL & One-Class Learning\\
    OCT & Optimal Classiciation Trees\\
    OCSVM & One-Class Support Vector Machine\\
    ODE & Ordinary Differential Equation\\
    OOB & Out-Of-Bag\\
    OUOB & Oversampling and Undersampling Online Bagging\\
    PILCO & Probablistic Inference for Learning Control\\
    PKnI & Prior Knowledge Integration\\
    PPO & Proximal Policy Optimization\\
    PRISMA & Preferred Reporting Item for Systematic Reviews and Meta-Analyses\\
    PRNN & Predictive Recurrent Neural Network\\
    PUL & Positive and Unlabeled Learning\\
    PU-LP & Positive and Unlabeled Learning by Label Propagation\\
    QL & Q-Learning\\
    RC-SVM & Rocchio Support Vector Machine\\
    RL & Reinforcement Learning\\
    RMTPP & Recurrent Marked Temporal Point Process\\
    ROSE & Robust Online Self-adjusting Ensemble\\
    RS & Reward Shaping\\
    RSF & Random Survival Forest\\
    SAT & Satisfiability problem\\
    SAMBA & Safe Model-Based and Active Reinforcement Learning\\
    SC & Spectral Clustering\\
    SC2 & StarCraft II\\
    SC2IL & StarCraft II Imitation Learning\\
    SC2LE & StarCraft II Learning Environment\\
    SDA & Spatial Depth Attention\\
    SFM & Social Force Modeling\\
    SGD & Stochastic Gradient Descent\\
    SLRR & Scalable Low-Rank Representation\\
    SMMC & Shooting Method Monte Carlo\\
    SMT & Satisfiability Modulo Theory \\
    SODEN & Survival model through Ordinary Differential Equation Networks\\
    SPP & Spatio-temporal Homogenous Poisson Process\\
    SPPO & Safety-constrained Proximal Policy Optimization\\
    SR & Super-Resolution\\
    SRGAN & Super-Resolution Generative Adversarial Network\\
    SRP & Streaming Random Patches\\
    SSL & Semi-Supervised Learning\\
    ST-RSF & Self-Trained Random Survival Forest\\
    ST-RSF+CCT & Self-Trained Random Survival Forest Corrected with Censored Times\\
    SVM & Support Vector Machine\\
    TC & Total Cost\\
    TELL & inTerpretable nEuraL cLustering\\
    TTC & Time-To-Conflict\\
    UMAP & Uniform Manifold Approximation and Projection\\
    VDSR & Very Deep Super Resolution Network\\
    WAIL & Wasserstein Arversarial Imitiation Learning\\
    XAI & Explainable Artificial Intelligence\\
    \bottomrule
\end{longtable}
\end{center}

\end{document}